\documentclass[letterpaper, 10 pt, conference]{ieeeconf}  

\IEEEoverridecommandlockouts                              

\usepackage{url}
\usepackage{graphicx}
\usepackage{amsmath}

\title{\LARGE \bf
Accelerometry-based Energy Expenditure Estimation During Activities of Daily Living: A Comparison Among Different Accelerometer Compositions
}

\author{Shuhao Que*$^{1}$, Remco Poelarends*$^{2}$, Peter Veltink$^{1}$, Miriam Vollenbroek-Hutten$^{1}$, and Ying Wang$^{1}$
\thanks{*These authors contributed equally to this work.}
\thanks{This study was approved by the ethical review board of University of Twente and funded by the HealthyW8 project.}
\thanks{$^{1}$S. Que, P. Veltink, M. Vollenbroek-Hutten, and Y. Wang are with the Department of Electrical Engineering, University of Twente, Enschede, The Netherlands.
        {\tt\small s.que@utwente.nl; \tt\small p.h.veltink@utwente.nl; \tt\small m.m.r.hutten@utwente.nl; \tt\small ying.wang@utwente.nl}}%
\thanks{$^{2}$R. Poelarends is with the Department of Nuclear Medicine, Isala, Zwolle, The Netherlands.
        {\tt\small r.j.poelarends@isala.nl}}%
}

\begin{document}
\maketitle
\thispagestyle{empty}
\pagestyle{empty}

\begin{abstract}
Physical activity energy expenditure (PAEE) is an important parameter to monitor as it can be potentially used for obesity control. PAEE can be measured from breath-by-breath respiratory data, which can serve as a reference. Alternatively, PAEE can be predicted from the body movements, which can be measured and estimated with accelerometers. The body center of mass (COM) acceleration reflects the movements of the whole body and thus serves as a good predictor for PAEE. Therefore, it is desirable to attach the accelerometers to locations close to the COM. However, the wrist has also become a popular location due to recent advancements in wrist-worn devices. Therefore, in this work, using the respiratory data measured by COSMED K5 as the reference, we evaluated and compared the performances of COM-based settings and wrist-based settings. The COM-based settings include two different accelerometer compositions, using only the pelvis accelerometer (pelvis-acc) and the pelvis accelerometer with two accelerometers from two thighs (3-acc). The wrist-based settings include using only the left wrist accelerometer (l-wrist-acc) and only the right wrist accelerometer (r-wrist-acc). We implemented two existing PAEE estimation methods on our collected dataset, where 9 participants performed activities of daily living while wearing 5 accelerometers (i.e., pelvis, two thighs, and two wrists). These two methods include a linear regression (LR) model and a CNN-LSTM model. Both models yielded the best results with the COM-based 3-acc setting (LR: $R^2$ = 0.41, CNN-LSTM: $R^2$ = 0.53). No significant difference was found between the 3-acc and pelvis-acc settings (p-value = 0.278). For both models, neither the l-wrist-acc nor the r-wrist-acc settings demonstrated predictive power on PAEE with $R^2$ values close to 0, significantly outperformed by the two COM-based settings (p-values $<$ 0.05). No significant difference was found between the two wrists (p-value = 0.329).
\end{abstract}

\section{INTRODUCTION}
Total daily energy expenditure (TEE) encompasses the sum of energy expended by the body during 24 hours, reflecting the energy expenditure of three components: basal metabolic rate, diet-induced thermogenesis, and physical activity energy expenditure (PAEE). Among the three components, PAEE showcases the strongest inter-subject variability (among individuals) and intra-subject variability (within a day)~\cite{b8}. It is also closely related to obesity control~\cite{westerterp2017control}. Thus, it is of substantial research interest to measure and gain insight into individuals' PAEE. Conventionally, based on the indirect calorimetry method~\cite{ferrannini1988theoretical}, PAEE can be calculated by subtracting measured BMR and DIT from TEE. However, indirect calorimetry is not sufficiently portable for daily life monitoring~\cite{b9}. As a result, there has been an increased focus on the development of wearable sensor-based monitoring tools for PAEE. The used wearable sensors usually incorporate one or multiple accelerometers. The systematic review by Jeran et al.~\cite{b11} revealed a variance in activity-related energy expenditure explained by accelerometry-assessed physical activity ranging from $R^2$ = 0.04 to $R^2$ = 0.80 (median crude $R^2$ = 0.26) under free-living conditions.

Regarding the accelerometer's sensor location for estimating PAEE, it is common practice to use a location close to the body center of mass (COM), such as the lower back~\cite{bouten1994assessment}. COM acceleration reflects movements of the whole body~\cite{farris1987prototype, smidt1971accelerographic} and is thus good for estimating PAEE. However, due to the development and availability of wrist-worn devices, such as Garmin, Apple Watch, Samsung Watch, Fitbit, and so on, the wrist has become a popular sensor location being investigated for estimating PAEE. Nevertheless, their usability in estimating PAEE is still questioned and needs further investigation~\cite{duking2020wrist, fuller2020reliability}. Altini et al.~\cite{altini2014estimating} compared 5 sensor locations (i.e., chest, dominant ankle, dominant thigh, dominant wrist, and right hip) individually and their compositions. They found that based on a single-sensor approach, the wrist model performed the worst while the chest model performed the best. They concluded that using a single sensor close to COM (chest or right hip) was sufficient for accurate PAEE estimation. However, the sensor location compositions were not selected to approximate the COM acceleration. Because of that, they also did not explicitly propose a model based on COM and compare it with the wrist model.

Is a single sensor sufficient to represent the COM acceleration? Does introducing more sensors improve the COM acceleration representation, thereby improving the PAEE estimation? What is the performance difference between a wrist-based setting and a COM-based setting? To address these questions, in this work, we validated two COM-based settings and two wrist-based settings for their usability of PAEE estimation during activities of daily living (ADL). We performed such validation on our collected dataset. Furthermore, we explicitly compared the performance of the COM-based settings and the wrist-based settings.  

\section{Methods}

\subsection{Data collection}
The study was conducted at the eHealth House, University of Twente. A total of 10 participants (30\% female) were recruited for this study. Inclusion criteria were: (1) aged between 18 and 60~\cite{pontzer2021daily}; (2) have a Body Mass Index (BMI) lower than $40 kg/m^2$ because according to Westerterp~\cite{westerterp2017control}, when one reaches the class III obese state (i.e., BMI $> 40 kg/m^2$) there is no further increase of PAEE; (3) free of cardiovascular diseases, respiratory diseases, metabolic disorders; (4) not being pregnant; (5) free of physical disabilities that impact daily living. The study was ethically approved by the Ethics Committee of Computer \& Information Science of the University of Twente (File no. 230728). Informed consent was obtained from all individual participants included in the study. Static data including age, sex, height, and weight, were collected. Body composition was estimated using a Bioelectrical Impedance Analysis (BIA) scale (Omron BF511). Information about recent physical activity levels was collected using the International Physical Activity Questionnaire (IPAQ)~\cite{brazier1992validating, craig2003international}.

\begin{table}[h]
\caption{List of activities with their corresponding duration. x represents variable activity duration.}
\label{tab:activities}
\begin{center}
\begin{tabular}{|c||c||c|}
\hline
Activity & Duration [s]\\
\hline
Sitting resting & 300 \\
Sitting reading & 300 \\
Standing still & 180 \\
Working on a laptop & x \\
Emptying dishwasher & x \\
Mopping & x \\
Stacking shelves with books & x \\
Climbing stairs (5 times) & x \\
Treadmill (3 km/h) & 300 \\
Treadmill (5 km/h) & 300 \\
Cycle at 125 Watt & 300 \\
\hline
\end{tabular}
\end{center}
\end{table}

During the data collection session, each participant started with a 30-minute quiet rest in the supine position to estimate rest metabolic rate (RMR)~\cite{nosslinger2021underestimation}. A series of ADL followed this, as presented in Table \ref{tab:activities}. Most activities were performed for at least 5 minutes to reach steady-state energy expenditure, as recommended by \cite{alvarez2020survey}. Each participant performed the activities in a randomized order to prevent the introduction of bias in the dataset~\cite{b12}.

\begin{figure}[!h]
    \centering
    \includegraphics[width=0.5\linewidth]{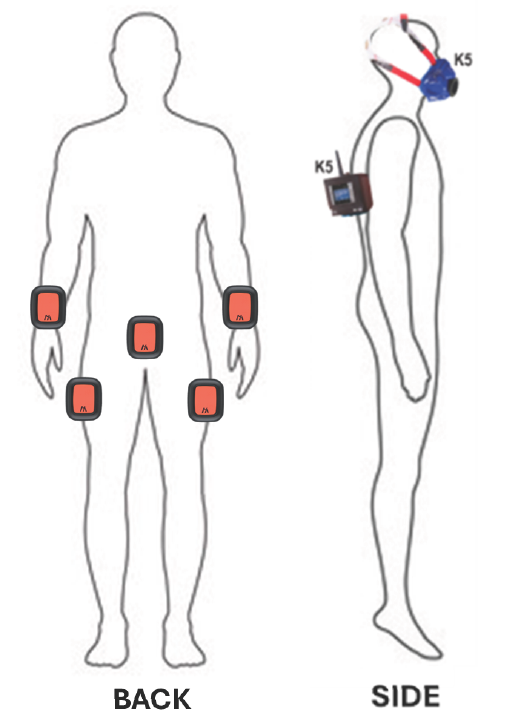}
    \caption{Distribution of the used sensors, including five Movella Xsens DOT sensors for the inertial measurements and COSMED K5 for the breath-by-breath respiratory data.}
    \label{fig:sensors}
\end{figure}

Objective data was captured using different wearable sensors. Three-axial acceleration data was collected at 30 Hz at five body locations, i.e., left wrist, right wrist, left thigh, right thigh, and pelvis (Movella Xsens DOT~\cite{movelladot}). Breath-by-breath respiratory data, serving as ground truth, was collected (COSMED K5~\cite{deblois2021reliability}) and activity labels were added using the OMNIA COSMED software. The sensor distribution is shown in Figure \ref{fig:sensors}.

\subsection{Data preprocessing}
To estimate the RMR in terms of the consumed oxygen flow ($V_{O_2}$, mL/min) and the produced carbon dioxide flow ($V_{CO_2}$, mL/min), the initial 5 minutes of the measurement during the quiet rest period were excluded to eliminate any transient effects from the onset of the resting
period. The average $V_{O_2}$ and $V_{CO_2}$ were calculated from the remaining data, as instructed by Compher et al.~\cite{compher2006best}, which were considered the participant’s RMR. The measured TEE during the ADL consists of the components PAEE and RMR. The average RMR $V_{O_2}$ and $V_{CO_2}$ were subtracted from the TEE, leading to the preprocessed $V_{O_2}$ and $V_{CO_2}$ measurements during the ADL. PAEE was then derived
using PAEE = TEE - RMR (in W/kg). The breath-by-breath $V_{O_2}$ and $V_{CO_2}$ data were first resampled to 1 Hz. They were then smoothed using a first-order Savitzky-Golay filter with a 20-second window length to reduce noise and correct for artifacts caused by talking. 

The collected raw acceleration data were corrected for gravitational acceleration to obtain gravity-free sensor acceleration. Subsequently, a Butterworth fourth-order lowpass filter with a cutoff frequency of 6 Hz was utilized to filter the acceleration data as done by Lee et al.~\cite{lee2024imu} for their energy expenditure model based on neural networks. Finally, all signals were resampled to 1 Hz and synchronized. 

\subsection{Models}
Linear regression (LR) and neural networks (NN) are two well-known types of models to estimate PAEE~\cite{b11}. Therefore, for this work, two existing models for accelerometry-based PAEE estimation were implemented, including the classic LR model by Boutan et al.~\cite{bouten1994assessment} and a Convolutional Neural Network and Long-Short-Term Memory (CNN-LSTM) model by Lee et al.~\cite{lee2024imu}. The LR method by Bouten et al.~\cite{bouten1994assessment} was selected because it was one of the classic works on using body acceleration to estimate PAEE. The CNN-LSTM model by Lee et al.~\cite{lee2024imu} was selected because their model was one of the latest works on using NN and accelerometers to estimate PAEE. In addition, their proposed CNN-LSTM model takes the pre-processed three-axial acceleration data as input, eliminating the need for manual feature engineering. 

For the model’s input, four accelerometer compositions were used, including two COM-based settings: only the pelvis accelerometer (pelvis-acc) and the pelvis accelerometer and two thigh accelerometers (3-acc), two wrist-based settings: only the left wrist accelerometer (l-wrist-acc), and only the right wrist accelerometer (r-wrist-acc).

\subsubsection{Model's input from a single accelerometer (pelvis-acc, l-wrist-acc, r-wrist-acc)} The accelerometry-derived feature $IAA_{tot}$ was used as input for the LR model. $IAA_{tot}$ is defined as the sum of the integrated absolute acceleration along three axes and was reported by Bouten et al.~\cite{bouten1994assessment} to be the most correlated with total energy expenditure in their dataset. Considering the impact of different accelerometer's axial positioning on the human body, $IAA_{tot}$ is sufficiently informative as it encodes information from three axes and thus serves as a robust feature against different sensor orientations. Its mathematical definition is given below:

\begin{align}
    IAA_{tot} = \sum_i^T |a_x[i]| + \sum_i^T |a_y[i]| + \sum_i^T |a_z[i]| \\
\end{align}

where $i$ denotes the $i$-th time point and $T$ denotes the total time points for integration. $|a_x|$, $|a_y|$, and $|a_z|$ denote the absolute acceleration along the x-axis, y-axis, and z-axis, respectively.

Both models were trained to perform 1-second horizon forecasting of the PAEE value, using the past 30-second accelerometer measurements. For the LR model, each 30-second sliding window with a step size equal to 1 second was processed to generate one $IAA_{tot}$ value. For the CNN-LSTM model, the preprocessed three-axial acceleration data were segmented by the 30-second sliding windows and used as the model input. The model input matrix has the size of 3-by-30, where 3 denotes the feature dimension (i.e., the 3 axes) and 30 denotes the length of the sliding window (i.e., 30 seconds). 

\subsubsection{Model's input from 3 accelerometers (3-acc)} For the LR model, we used three regressors (i.e., independent variables), which indicated three $IAA_{tot}$ calculated from each accelerometer signal (e.g., PAEE = a*$IAA_{tot_1}$ + b*$IAA_{tot_2}$ + c*$IAA_{tot_3}$ + d). For the CNN-LSTM model, the three-axial acceleration data from each accelerometer was concatenated along the feature dimension and used as the input matrix with the size of 9-by-30.

\subsection{Evaluation metrics}
Leave-one-subject-out cross validation was used to evaluate the models' performance. The normalized root-mean-squared error (NRMSE) was used to evaluate the models' predicted PAEE compared with the ground truth:

\begin{equation}
    NRMSE = \frac{\sqrt{\sum^n_{i=1} \frac{(y_i - x_i)^2}{n}}}{\bar{y}}
\end{equation}

where n represents the total number of data points. $y_i$ and $x_i$ represent the ground truth and the predicted PAEE at time index $i$, respectively. $\bar{y}$ represents the mean ground truth value of PAEE. 

The R-squared values were calculated to evaluate the models' capacity to capture the variations in PAEE due to ADL. It is defined as:

\begin{equation}
    R^2 = 1 - \frac{\sum_i (x_i - y_i)^2}{\sum_i (x_i - \bar{y})^2}
\end{equation}

The repeated measures ANOVA test~\cite{rutherford2011anova} was used to evaluate the statistical significance (p-value $<$ 0.05) of the mean difference among the four different sensor location compositions of the two models (LR and CNN-LSTM).
The two-sided Student's paired t-test~\cite{student1908probable} was used for pairwise comparisons, with Bonferroni correction~\cite{bonferroni1936teoria} applied to adjust for multiple comparisons. The Shapiro-Wilk test~\cite{shapiro1965analysis} was used to validate the normal distribution assumption (p-value $>$ 0.05) required by both the repeated measures ANOVA test and paired t-test.

\section{Results and Discussions}
In total, 9 participants' data were used for further analysis because data from 1 participant was incomplete and excluded due to a loss of Bluetooth connection. 

The results achieved by the LR model and the CNN-LSTM model are summarized in Table \ref{tab:resall}. The NRMSE and $R^2$ values are all normally distributed according to the Shapiro-Wilk test (p-values $>$ 0.05). According to the repeated measures ANOVA test, no significant difference was found between the two models (NRMSE p-value = 0.164, $R^2$ p-value = 0.253). However, significant differences were found among different sensor location compositions (NRMSE p-value = 0.000, $R^2$ p-value = 0.000). According to the Student's paired t-test for post-hoc pairwise comparisons, no significant difference was found between the 3-acc and the pelvis-acc, with Bonferroni-corrected p-values of 0.269 and 0.278 for NRMSE and $R^2$, respectively. The two COM-based settings (3-acc and pelvis-acc) significantly outperformed the wrist-based settings (p-values $<$ 0.05). Both wrists exhibited almost no predictive power on PAEE, with $R^2$ values close to 0. No significant difference was found between the left and right wrists (NRMSE p-value = 0.283, $R^2$ p-value = 0.329). 

\begin{table}[!h]
\caption{Results of both linear regression (LR) model and the CNN-LSTM model. SD is short for standard deviation. The respective best results of the LR and CNN-LSTM models are in bold font.}
    \centering
    \begin{tabular}{|c|c|c|c|}
        \hline
         Accelerometer  & Model & NRMSE & $R^2$ \\
         Composition & & Mean (SD) & Mean (SD)\\
         \hline
         pelvis-acc & LR & 0.88 (0.21) & 0.18 (0.20) \\
         (pelvis) & & &\\
        \hline
         3-acc & LR & \textbf{0.72 (0.22)} & \textbf{0.41 (0.26)} \\
         (pelvis, two thighs) & & &\\
        \hline
         r-wrist-acc & LR & 0.95 (0.19) & 0.03 (0.20) \\
         (right wrist) & & &\\
         \hline
         l-wrist-acc & LR & 0.94 (0.20) & 0.06 (0.21) \\
         (left wrist) & & & \\
         \hline
         pelvis-acc  & CNN-LSTM & 0.75 (0.19) & 0.38 (0.23)\\
         (pelvis) & & & \\
        \hline
         3-acc & CNN-LSTM & \textbf{0.64 (0.18)} & \textbf{0.53 (0.26)}\\
         (pelvis, two thighs) & & &\\
        \hline
         r-wrist-acc & CNN-LSTM & 0.97 (0.14) & -0.02 (0.10) \\
        (right wrist) & & &\\
        \hline
         l-wrist-acc & CNN-LSTM & 0.95 (0.16) & 0.03 (0.11) \\
        (right wrist) & & &\\
         \hline
    \end{tabular}
    \label{tab:resall}
\end{table}

\begin{figure*}[!h]
    \centering
    \includegraphics[width=\linewidth]{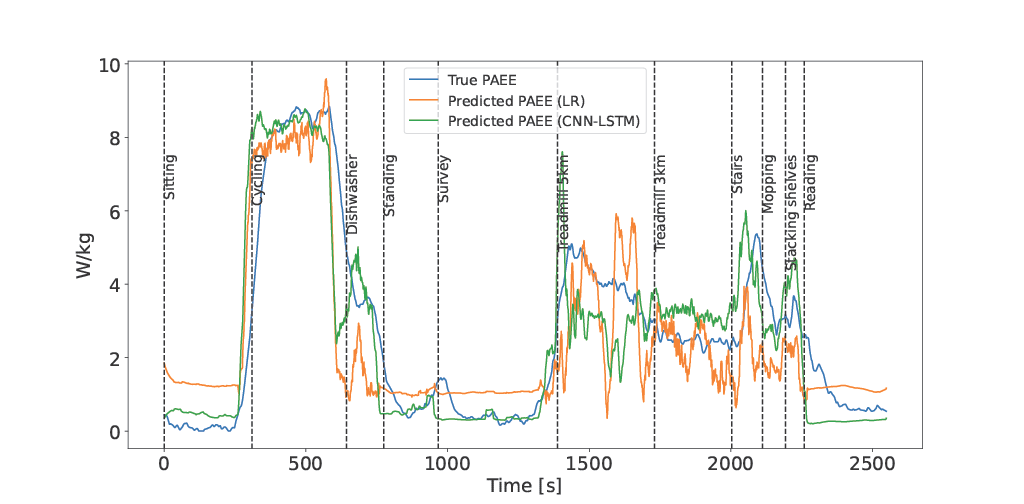}
    \caption{Good estimation results (3-acc) on one test subject by the LR model (NRMSE = 0.517, $R^2$ = 0.691) and CNN-LSTM model (NRMSE = 0.460, $R^2$ = 0.755). For the LR model, estimated PAEE = -0.058 * $IAA_{tot_1}$ + 0.053 * $IAA_{tot_2}$ + 0.001 * $IAA_{tot_3}$ + 0.906, where the three $IAA_{tot}$ features were extracted from the pelvis, left thigh, and right thigh, respectively.}
    \label{fig:eegood}
\end{figure*}

\begin{figure*}[!h]
    \centering
    \includegraphics[width=\linewidth]{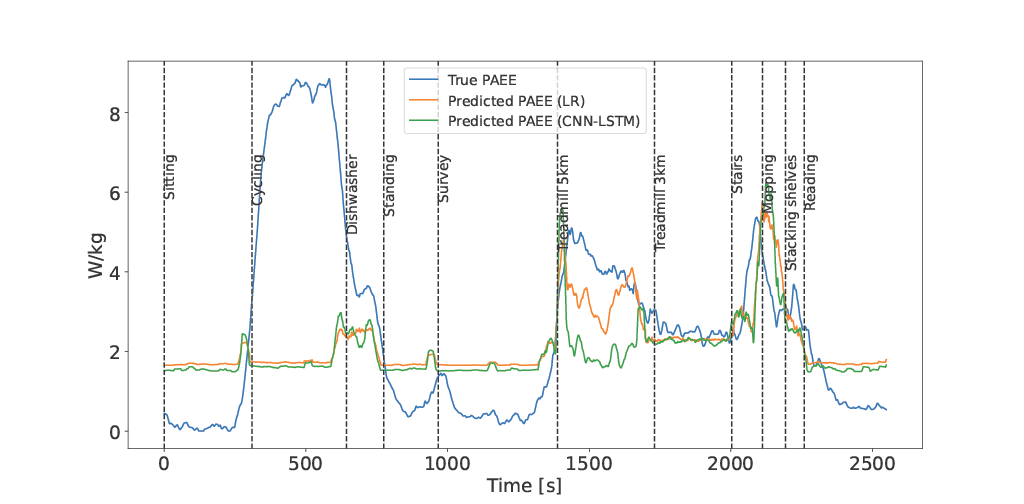}
    \caption{Bad estimation results (r-wrist-acc) on one test subject by the LR model (NRMSE = 0.911, $R^2$ = 0.040) and CNN-LSTM model (NRMSE = 0.971, $R^2$ = -0.091). For the LR model, estimated PAEE = 0.011 * $IAA_{tot}$ + 1.632, where the $IAA_{tot}$ feature was extracted from the right wrist.}
    \label{fig:eebad}
\end{figure*}

An example of good estimation on one test subject by both models in the time domain is shown in Figure \ref{fig:eegood}. As observed, for this test subject, both models in the 3-acc setting captured the general PAEE changes over the course of ADL. Using the r-wrist-acc setting, an example of bad estimation on the same test subject is shown in Figure \ref{fig:eebad}. In this case, both models underestimated the PAEE values during cycling activity. In addition, both models' predicted baseline PAEE values are close to 2 W/kg instead of close to 0 W/kg, overestimating during the sitting, standing, survey, and reading activities.

Using the COM-based settings, both models yielded decent results in predicting PAEE with no significant difference between one another. This showcased the strong predictive power of the feature $IAA_{tot}$ using LR, comparable to CNN-LSTM. This also indicated the potential correlation between body acceleration integrated over time and PAEE. 

No significant difference was found between the two COM-based settings: 3-acc and pelvis-acc. This suggests that using a single sensor is sufficient to represent the COM acceleration. The left wrist and right wrist demonstrated limited to no predictive power over PAEE, which is likely due to their poor reflection of the COM acceleration. Although existing commercial wrist-worn devices have claimed their usability of tracking users’ PAEE, our results indicate that such claims should be interpreted with caution especially when only wrist-worn acceleration information is used to estimate PAEE. With additional sensor information, like heart rate, acquired from wrist-worn devices, the PAEE performance might improve. However, considering the faraway location of the wrist to the COM and its poor performance tested in this study, we highly recommend further validation and development for wrist-based PAEE estimation in daily health monitoring.

In future work, a larger dataset should be used to further validate the NN-based method and re-evaluate statistical significance. Additionally, more accelerometer configurations (e.g., chest or upper arm) should be explored to better represent or even derive COM acceleration. To fully leverage wrist-worn sensor technology, future research should investigate methods to ensure that wrist acceleration accurately reflects COM acceleration before using it to estimate PAEE.

\section{Conclusions}
Based on two well-known modeling approaches (LR and NN), the wrist-based settings demonstrated almost no predictive power on PAEE while the subject performed a sequence of ADL. The COM-based settings significantly outperformed the wrist-based settings. For accurate PAEE estimation, we suggest following the COM-based modeling approach, and the number and location of the deployed accelerometer be determined based on how well they reflect the COM acceleration.

\addtolength{\textheight}{-12cm}   

\end{document}